# Evaluation of the Accuracy of the BGLemmatizer

**Elena Karashtranova, Grigor Iliev, Nadezhda Borisova,
Yana Chankova, Irena Atanasova**

*South-West University, Blagoevgrad, Bulgaria*

***Abstract:*** *This paper reveals the results of an analysis of the accuracy of developed software for automatic lemmatization for the Bulgarian language. This lemmatization software is written entirely in Java and is distributed as a GATE plugin. Certain statistical methods are used to define the accuracy of this software. The results of the analysis show 95% lemmatization accuracy.*

***Keywords:*** *Bulgarian grammar, NLP, GATE*

## 1. INTRODUCTION

The entry for "lemmatize" in the *Collins English Dictionary* states that lemmatization is the process of grouping together the different inflected forms of a word so that they can be analyzed as a single term (Collins English Dictionary[1]). Lemmatization is a fundamental natural language processing (NLP) task which automates the process of word normalization. The correct identification of the normalized form of a word is of significance for NLP tasks such as information extraction and information retrieval. It becomes of particular interest when applied to highly inflectional languages such as Bulgarian, which is a South Slavic language.

The total amount of relevant information in a sample is controlled by two factors:

- The sampling plan or experimental design which represents the procedure for collecting the information;
- The sample size *n* or the amount of information one collects [4].

The corpus we have investigated consists of 273933 tokens [1]. Every token is manually annotated with part-of-speech tags [5]. The aim of our study is to evaluate the lemmatizer's performance regarding three parts of speech, namely the noun, the adjective and the verb. In order to improve the accuracy of analysis results we have investigated only these parts of speech as a population with size 149061. We have conducted the survey in two stages (i.e. we have examined a two-phase sampling). Our discussion begins with an analysis of the results of a pilot study as such an approach has two advantages. First, the pilot study will be used to provide estimates of the individual stratum variances and second, the results of the pilot study can be used to estimate the number of observations needed to obtain estimators of the population parameters with a specified level of precision.

---

[1] http://www.collinsdictionary.com

On the basis of the results obtained from the samples, the traditional evaluation metrics have been applied, namely, **Precision, Recall** and **F-measure** [2].

**Precision**, as is well-known, measures the number of the items that have been correctly identified as a percentage of the number of the identified items. The higher the **Precision** is, the better the system is at ensuring what has been identified as being correct.

**Recall** measures the number of the items that have been correctly identified as a percentage of the total number of items. The higher the **Recall** rate is, the better the system is at not missing correct items.

The **F$_β$-measure** is used together with **Precision** and **Recall,** as a weighted average of both **Precision** and **Recall**.

## 2. SURVEY

As mentioned above, we have conducted a pilot study, sampling 40 observations from each district, whereby the numbers in the three stratum district are [4]:

$N_1$=80509 /for the nouns/
$N_2$=23159 /for the adjectives/
$N_3$=45393 /for the verbs/

The numbers $n_1, n_2, n_3$ have been sampled in the three strata as follows:

(1) $\quad n_j = \frac{N_j \sigma_j}{\sum_{i=1}^{3} N_i \sigma_i} \cdot n \qquad$ *(j=1,2,3),*

where $\sigma_i$ marks the stratum population standard deviations. The sample stratum standard deviations obtained are $\sigma_1$=0,243, $\sigma_2$=0,352, $\sigma_3$=0,195.

By substituting our sample estimates in place of quantities (1), we have found out that:

$n_1$=0,534.n; $n_2$=0,224.n; $n_3$=0,242.n

We have now specified the proportions of the total sample to be allocated to each stratum under the optimal scheme.

By means of (2) we can find the total number of the sample:

(2) $\quad n = \frac{\frac{1}{N}\left(\sum_{i=1}^{3} N_j \sigma_j\right)^2}{N\sigma_{\hat{p}}^2 + \frac{1}{N}\sum_{j=1}^{K} N_j \sigma_j^2}$

where N=149061 is the total number of the population members and $\sigma_{\hat{p}}^2$ is the variance of the estimator of the population proportion.

In order that the 95% confidence interval for the population proportion be achieved we extend the error margin by 0,02 on each side of the sample estimate $(\sigma_{\hat{p}} = 0{,}02)$.

Hence, we can conclude that the needed total number of sample observations is 597.

Given that it is easy to make a random sample, the total number of sample observations amounted to 1373. These have then been allocated among the three strata as follows:

$n_1 = 0{,}534 \cdot 1373 = 732$
$n_2 = 0{,}224 \cdot 1373 = 308$
$n_3 = 0{,}242 \cdot 1373 = 333$

Since 40 observations have already been sampled in each stratum, the numbers sampled in the second phase are 692, 268, 293 respectively.

We have estimated the sample proportion by means of (3):

(3)    $\hat{p} = \dfrac{\sum_1^3 n_i \cdot p_i}{\sum_1^3 n_i},$

where $p_i$ marks the proportions of the investigated parameters in each stratum and $n_i$ marks the numbers of the sampled stratum [3].

In view of **Precision** we resave $\hat{P} = 0{,}97$ and 95% confidence interval for the population **Precision** is:

$$0{,}97 - 0{,}02 < P < 0{,}97 + 0{,}02$$

Concerning **Recall** we resave $\hat{R} = 0{,}93$ and *95%* confidence interval for the population **Recall** is:

$$0{,}93 - 0{,}02 < R < 0{,}93 + 0{,}02$$

In our study the **F-measure** is used as a weighted average of both **Precision** and **Recall** which are considered as equally important, so that:

$F = \dfrac{P \cdot R}{0{,}5(P+R)} = 0{,}95$

Such results are highly satisfactory and bear out the high accuracy and precision of the developed lemmatizer.

For the purpose of the following analyses, we have designed the frequency distribution on the basis of the specific features of each part of speech.

In order to achieve greater objectivity of verification of the sample observations, the parts of speech are retrieved with context. The specifics of our study design facilitate the procedure of eliciting extensive information on the structure of the different parts of speech included in the corpus. In view of

Mathematics and Informatics

the above, we have built the necessary frequency distributions which are demonstrated in the following tables.

Table 1. Frequency distribution of the Nouns

| BTB-TS tag | Frequency |
|---|---|
| N-msi | 18667 |
| N-msh | 4918 |
| N-msf | 2560 |
| N-mpi | 5004 |
| N-mpd | 3136 |
| N-mt | 1966 |
| N-fsi | 15816 |
| N-fsd | 6127 |
| N-fpi | 4992 |
| N-fpd | 1836 |
| N-nsi | 7398 |
| N-nsd | 4288 |
| N-npi | 1992 |
| N-npd | 986 |

Table 2. Frequency distribution of the Adjectives

| BTB-TS tag | Frequency |
|---|---|
| Amsi | 3256 |
| Amsh | 2062 |
| Amsf | 1099 |
| Afsi | 3287 |
| Afsd | 2785 |
| Ansi | 2074 |
| Ansd | 1492 |
| A-pi | 4172 |
| A-pd | 2811 |

Table 3. Frequency distribution of the Verbs

| BTB-TS tag | Frequency | BTB-TS tag | Frequency | BTB-TS tag | Frequency |
|---|---|---|---|---|---|
| V---f-r1s | 1769 | V---u-o2p | 20 | V---cv—sfi | 669 |
| V---f-r2s | 656 | V---u-o3p | 22 | V---cv—sfd | 130 |
| V---f-r3s | 15582 | V---z--2s | 217 | V---cv—sni | 522 |
| V---f-r1p | 1391 | V---z---p | 158 | V---cv—snd | 92 |
| V---f-r2p | 634 | V---cao-smi | 1204 | V---cv--p-i | 1305 |
| V---f-r3p | 6711 | V---cao-smh | 55 | V---cv--p-d | 327 |
| V---f-t1s | 42 | V---cao-smf | 3 | V---car-smi | 74 |
| V---f-t2s | 4 | V---cao-sfi | 478 | V---car-smh | 46 |
| V---f-t3s | 831 | V---cao-sfd | 120 | V---car-smf | 17 |
| V---f-t1p | 11 | V---cao-sni | 314 | V---car-sfi | 63 |
| V---f-t2p | 7 | V---cao-snd | 25 | V---car-sfd | 107 |
| V---f-t3p | 259 | V---cao-p-i | 941 | V---car-sni | 50 |
| V---u-o1s | 53 | V---cao-p-d | 133 | V---car-snd | 48 |
| V---u-o2s | 3 | V---cv—smi | 1124 | V---car-p-i | 165 |
| V---u-o3s | 112 | V---cv—smh | 96 | V---car-p-d | 162 |
| V---u-o1p | 5 | V---cv—smf | 50 | | |

## 3. CONCLUSION

The above frequency distribution tables can be used in performing analysis of the errors and detecting the types of errors in view of their elimination which will contribute to increasing the precision and efficiency of the developed software and enhance the possibilities for its application in different NLP tasks, such as information extraction and information retrieval.

We suggest that the samples should be made randomly in keeping with the proportions presented in the frequency distributions. The proposed method of choosing the sample size can be used in performing the estimation procedure for the three cases listed above. The parameters that are to be estimated as well as the standard error margin are determined on the basis of their point estimator.

## 4. ACKNOWLEDGEMENTS

This research was funded by the South-West University "Neofit Rilski" grant SRP-A18/15.